\PassOptionsToPackage{table,dvipsnames}{xcolor}
\documentclass[letterpaper]{article} 
\usepackage{aaai2026}  
\usepackage{times}  
\usepackage{helvet}  
\usepackage{courier}  
\usepackage[hyphens]{url}  
\usepackage{graphicx} 
\usepackage{natbib}  
\usepackage{caption} 
\usepackage{algorithmicx}
\usepackage{algorithm}
\usepackage{algpascal}
\usepackage{algc}
\usepackage{algcompatible}
\usepackage{amsmath}
\usepackage{amssymb}
\usepackage[table]{xcolor} 
\usepackage{boldline, hhline}
\usepackage{colortbl}  
\usepackage{multirow}
\usepackage{makecell}
\usepackage{array}
\usepackage{xcolor}
\usepackage{booktabs}  

\usepackage{newfloat}
\usepackage{listings}
\DeclareCaptionStyle{ruled}{labelfont=normalfont,labelsep=colon,strut=off} 
\floatstyle{ruled}
\newfloat{listing}{tb}{lst}{}
\floatname{listing}{Listing}
\title{FedARKS: Federated Aggregation via Robust and Discriminative Knowledge Selection and Integration for Person Re-identification}
\newcommand{\corrauthor}{\thanks{Corresponding author.}}
\author{
    Xin Xu\textsuperscript{\rm 1},
    Binchang Ma\textsuperscript{\rm 1},
    Zhixi Yu\textsuperscript{\rm 1}\corrauthor,
    Wei Liu\textsuperscript{\rm 1}\footnotemark[1]
}
\affiliations{
    \textsuperscript{\rm 1} School of Computer Science and Technology, Wuhan University of Science and Technology, Wuhan 430065 China \\
    xuxin@wust.edu.cn, 202213407218@wust.edu.cn, yuzhixi@wust.edu.cn, liuwei@wust.edu.cn
}

\usepackage{bibentry}

\begin{document}

\maketitle

\begin{abstract}
The application of federated domain generalization in person re-identification (FedDG-ReID) aims to enhance the model's generalization ability in unseen domains while protecting client data privacy. However, existing mainstream methods typically rely on global feature representations and simple averaging operations for model aggregation, leading to two limitations in domain generalization: (1) Using only global features makes it difficult to capture subtle, domain-invariant local details (such as accessories or textures); (2) Uniform parameter averaging treats all clients as equivalent, ignoring their differences in robust feature extraction capabilities, thereby diluting the contributions of high-quality clients. To address these issues, we propose a novel federated learning framework—\textbf{Fed}erated \textbf{A}ggregation via \textbf{R}obust and Discriminative \textbf{K}nowledge \textbf{S}election and Integration (\textbf{FedARKS})—comprising two mechanisms: RK (Robust Knowledge) and KS (Knowledge Selection). In our design, each client employs a dual-branch network of RK: the Global Feature Processing Branch serves as the primary component, extracting overall representations for model aggregation and server-side updates; while the Body Part Processing Branch acts as an auxiliary component, focusing on extracting domain-invariant local details to supplement and guide the local training process during global feature learning. Additionally, our KS mechanism adaptively assigns corresponding aggregation weights to clients based on their ability to extract domain-invariant knowledge, enabling the server to better integrate cross-domain invariant knowledge extracted by clients. Extensive experiments validate that FedARKS achieves state-of-the-art generalization results on the FedDG-ReID benchmark, demonstrating that learning subtle body part features can effectively assist and reinforce global representations, thereby enabling robust cross-domain person ReID capabilities.
\end{abstract}



\section{Introduction}
Person Re-identification (ReID) aims to retrieve target pedestrians across non-overlapping camera views, playing a crucial role in smart city applications such as public security and missing person searches. Although deep learning methods\cite{He_2016_CVPR,8721151,10.1145/3240508.3240552,9336268,Sun_2018_ECCV}have significantly improved ReID performance, they remain highly susceptible to domain shift, resulting in substantial performance degradation and reduced reliability when deployed in new environments with different camera characteristics, lighting conditions, or pedestrian distributions.

Recent efforts~\cite{Zhao_2021_CVPR,Song_2019_CVPR,Dai_2021_CVPR}address this challenge through Domain Generalization (DG) techniques, which train models on multiple labeled source domains to learn domain-invariant representations and mitigate domain shift. However, growing concerns about personal data protection restrict the development of DG-ReID, as its training relies on centralizing large volumes of personal image data, posing potential privacy risks to individuals. Therefore, it is necessary to develop privacy-preserving DG-ReID methods.

\begin{figure}[t]
\centering
\includegraphics[width=0.52\textwidth,trim=5.5cm 2.5cm 3.5cm 1cm,clip]{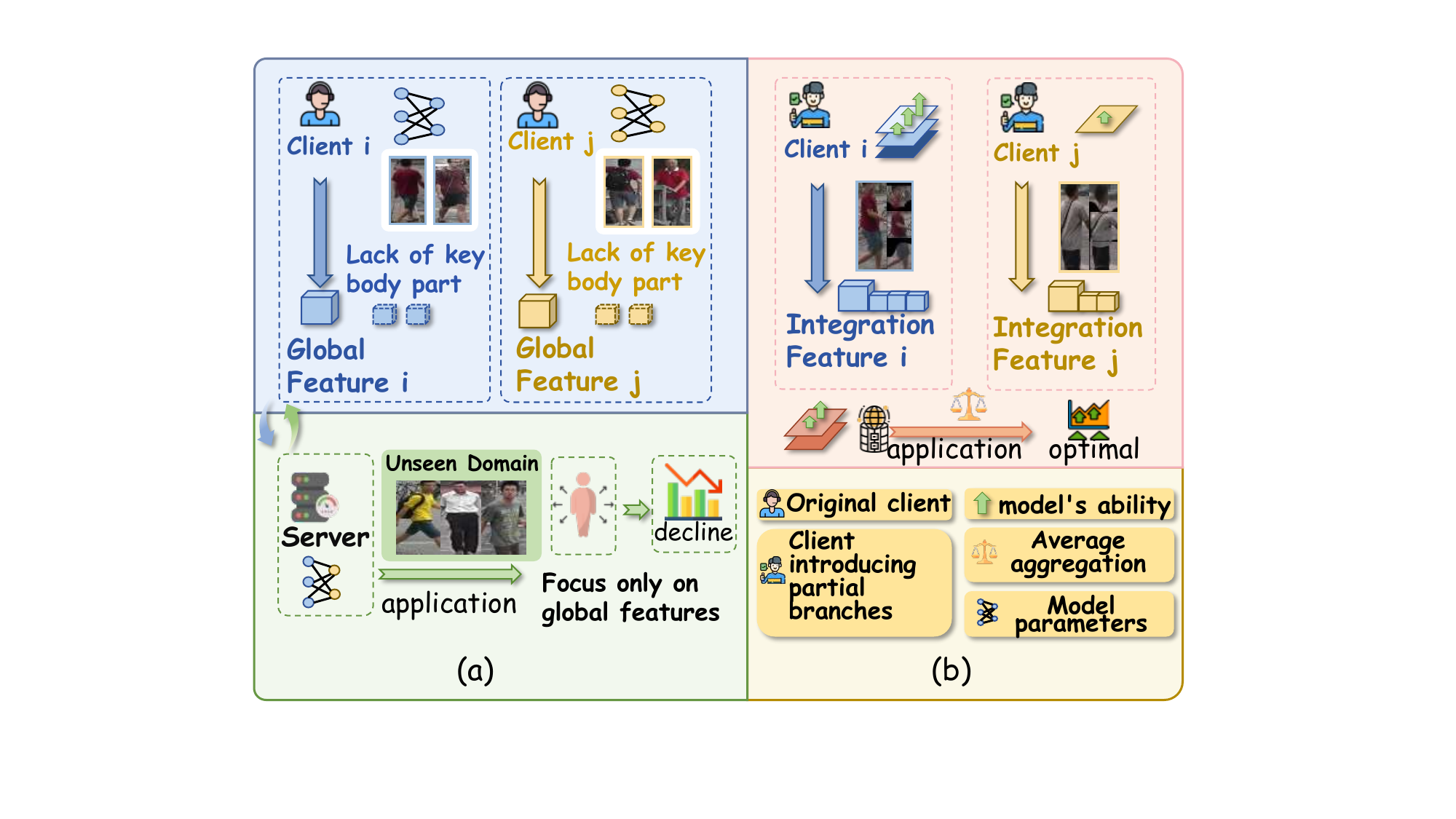}
\caption{Core challenges in FedDG-ReID: (a) Ignoring key local details that are invariant across domains; (b) Aggregation methods based on averaging dilute the ability of clients that are good at extracting cross-domain invariant features, thereby suppressing overall generalization ability.}
\label{fig1}
\end{figure}

To address privacy preservation concerns, recent research has applied Federated Learning (FL) to Domain Generalization in Person Re-identification (FedDG-ReID). While existing FedDG-ReID approaches\cite{wu2021decentralised}have made progress in mitigating local overfitting and adapting generic FL techniques, they crucially overlook two fundamental challenges:

\textbf{First}, as visualized in Fig. \ref{fig1}(a), cross-client \textbf{feature misalignment} arises when key local body part details—crucial for identity recognition and often possessing stronger cross-domain invariance—are neglected in feature representations. Specifically, due to data heterogeneity and limited coverage, clients often miss key body-part details critical for identity recognition and cross-domain invariance. Consequently, local models miss these important discriminative cues, leading to feature misalignment during aggregation, degrading generalization on unseen domains.

\textbf{Second}, as shown in Fig. \ref{fig1}(b), conventional averaging-based aggregation methods indiscriminately merge client updates, overlooking differences in each client’s capacity to capture robust, domain-invariant representations. As a result, the distinct advantages of clients adept at extracting such transferable features are diminished, leading to suboptimal integration of discriminative knowledge at the server. This ultimately suppresses the global model’s ability to generalize effectively across diverse domains.

\begin{figure*}[t]
\centering
\includegraphics[width=1.05\textwidth,trim=2.8cm 1cm 2cm 1cm,clip]{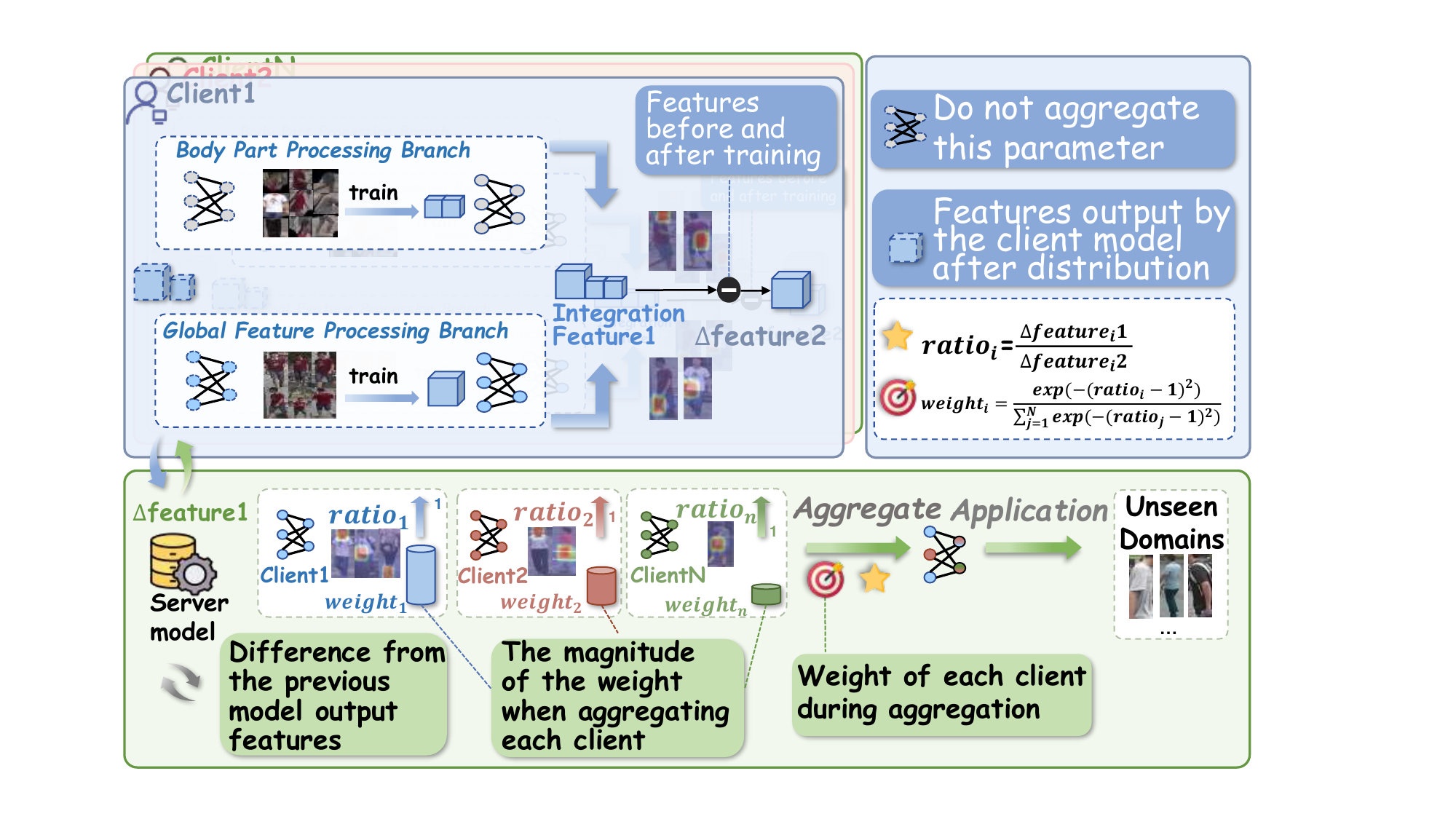}
\caption{Architectural diagram of FedARKS. Our method employs local branch training, in which each client fuses features in a dual-branch network. This figure illustrates the detailed process of local feature fusion and parameter aggregation, using client 1 as an example (other clients are shown in the background). After local training is complete, the server dynamically calibrates the aggregation weights of each client based on feature difference vectors and updates the global model in the next round of communication to improve cross-domain generalization capabilities.}
\label{fig2}
\end{figure*}

To address these challenges, we propose FedARKS (as shown in Fig.\ref{fig2}), which introduces two complementary mechanisms for enhanced domain generalization. The first mechanism, FedARKS-RK (Robust Knowledge), is deployed on each client and consists of two specialized branches: a body part processing branch for learning fine-grained local features, and a global feature processing branch for capturing holistic representations. To alleviate cross-client pedestrian feature misalignment, we fuse global features (from the Global Feature Processing Branch) and part-combined features (from the Body Part Processing Branch) via weighted averaging. Crucially, only the global branch parameters are aggregated, allowing the server to model domain-invariant features for better generalization. The part branch parameters remain local, where they can effectively capture client-specific, fine-grained cues based on local data. Aggregating these heterogeneous part parameters would undermine their discriminative power, adaptability and capacity for personalized representation

Building upon this, the second mechanism, FedARKS-KS (Knowledge Selection), is introduced to make the most of the robust local knowledge developed by FedARKS-RK.
Specifically, the KS module adaptively assigns greater aggregation weights to clients with stronger domain-invariant feature extraction capabilities. This targeted aggregation mechanism ensures that the server can fully leverage the domain-invariant knowledge from all clients, thus amplifying the overall generalization capacity of the global model on both seen and unseen domains.Comprehensive experiments validate the effectiveness of our framework, showing robust generalization across person ReID datasets. Our contributions are summarized as follows:
\begin{itemize}
\item We analyze the key factors impacting the generalization ability of federated person ReID models, revealing that traditional aggregation often overlooks subtle, domain-invariant local features, and fails to utilize the domain-generalization advancements contributed by individual clients, resulting in limited cross-domain robustness of the aggregated model.
\item We propose FedARKS, a federated learning framework specifically designed for person ReID, whose core lies in the combination of two complementary mechanisms. FedARKS combines robust local feature extraction with adaptive knowledge aggregation: at each client, it separates and retains fine-grained local feature representations with critical local details, while the server selectively integrates these cross-domain invariant features from all clients. This design enables the server-side model to fully integrate domain-invariant features, thereby enhancing its generalization capabilities.
\item We introduce the Representation Selection (RS) and Knowledge Selection (KS) modules. The RS module enables each client to capture subtle and highly discriminative local features with strong cross-domain invariance, whereas the KS module adaptively assigns higher aggregation weights to clients with superior domain-invariant representation learning capabilities. This collaborative mechanism ensures that the global model fully leverages transferable knowledge across diverse domains and mitigates the adverse effects of feature misalignment and indiscriminate aggregation.
\end{itemize}
\section{Related Work}
\subsection{Domain Generalization} Domain Generalization (DG), due to its significant practical value, has become an emerging hotspot in the field of AI. Current mainstream DG methods, such as domain alignment techniques\cite{Li_2018_CVPR,Li_2018_ECCV,li2018domain}, meta-learning frameworks\cite{Li_2019_ICCV,10.1007/978-3-030-58607-2_12,Zhao_2021_CVPR,li2018learning}, and style/data augmentation strategies\cite{10.1007/978-3-030-58517-4_33,Zhang_2022_CVPR}, have demonstrated excellent performance under centralized training settings.Existing domain generalization (DG) methods achieve strong performance in centralized settings, yet their frequent dependence on aggregated data collection poses substantial privacy concerns. Within federated learning (FL), DG encounters critical obstacles including client-level data scarcity – characterized by insufficient sample diversity – and the absence of domain-specific labels. These limitations severely hinder the deployment of meta-learning frameworks and domain alignment mechanisms. Notably, for challenging ReID applications, the ability to extract and distill fine-grained local details—such as body-part features—has proven crucial for domain-invariant representation, but remains challenging to leverage under federated constraints.
\\
\subsection{Federated Learning} Federated Learning (FL)\cite{konecny2016federated3,pmlr-v54-mcmahan17a,7958569,GGEUR_CVPR25,FLSurveyandBenchmarkforGenRobFair_TPAMI24,FCCLPlus_TPAMI23,FPL_CVPR23,FCCL_CVPR22,liu2025mix,xu2022towards} enables multiple clients to collaboratively train shared or personalized models while preserving data privacy. The early representative FedAvg\cite{pmlr-v54-mcmahan17a}adopts a mechanism of local model averaging and redistribution. Subsequent studies like FedProx\cite{MLSYS2020_1f5fe839}, MOON\cite{Li_2021_CVPR}, and SCAFFOLD\cite{pmlr-v119-karimireddy20a} aim to mitigate local overfitting issues, though they primarily target closed-set settings (e.g., image classification with fixed classes). To address open-set\cite{Busto_2017_ICCV} tasks, Zhuang et al.\cite{10.1145/3394171.3413814}proposed FedPav, which adapts to re-ID tasks by exchanging feature extractors. Wu et al.\cite{wu2021decentralised}pioneered the definition of the FedDG-ReID problem (where each source domain corresponds to a client) and introduced model distillation as a solution. The core of these methods revolves around maintaining consistency between local and global model.However, federated settings for person ReID must also contend with severe occlusions and intra-client variation, motivating the need for structured, body part-aware representations. Building upon this motivation, recent methods\cite{yang2022pafm,Gao_2020_CVPR,Miao_2019_ICCV,WANG2021104186,XU2021106554,9236648,Kalayeh_2018_CVPR,Song_2018_CVPR,s20164431} suggest reconstructing structured sets of body parts within the source domain, enabling networks to focus on visible, informative regions, even without universal part annotations.

Specifically, we design a dual-branch network architecture, where one branch specializes in learning and distilling feature knowledge from these predefined body parts, and the other branch serves as the primary model learning from source domain data for re-identification tasks. Through our RK mechanism, the fine-grained features from the body-part branch are effectively distilled into the primary model, thus greatly enhancing its discriminative power on general pedestrian datasets and significantly improving model generalization across diverse unseen domains, even when facing substantial domain shifts or novel environments during deployment. This work draws on the stylization method proposed by\cite{yang2024diversity}, while further demonstrating its effectiveness in real-world cross-domain scenarios.
\\
\section{Methodology}
\subsection{Preliminaries}
\textbf{Problem Definition.} Given N labeled source domains $S = \lbrace D_1, D_2, \dots, D_K \rbrace$, where the $i$-th domain $D_i= \lbrace X_i, Y_i\rbrace$ is
comprised of $M_i$ training images $X_i$ and their correspond-
ing identity labels $Y_i$. FedDG-ReID constructs a federated architecture where each data domain serves as an independent node, enforcing strict data isolation constraints. Its core principle lies in leveraging a central server to coordinate distributed clients. Through a Client-Server Collaborative Learning (CSCL) mechanism, it jointly optimizes the person ReID model to enhance its generalization performance on unseen target domains. This framework adheres to the cross-silo federated learning paradigm\cite{MAL-083}. Under the strict prohibition of data centralization, FedDG-ReID presents greater challenges compared to classical domain generalization methods for person ReID.
\subsection{Extraction of Robust Knowledge}
\textbf{Motivation.} In federated person re-identification, data collected from distributed clients exhibit significant heterogeneity due to variations in viewpoint, pose and background, presenting a non-IID challenge.Traditional global feature learning paradigms primarily emphasize holistic appearance cues derived from the entire person bounding box, often systematically overlooking subtle yet domain-invariant local discriminative features—such as distinctive accessories, unique fabric textures, or semantically key body part—which are crucial for robust matching under cross-client variations. This critical oversight inevitably results in suboptimal feature alignment during aggregation and severely hampers cross-client generalization. To decisively overcome these limitations, we propose integrating a dedicated RK module within each client's local learning process; this module is meticulously designed to extract specific body parts (e.g., head/shoulders, torso, lower body/legs) .This module enables (1) robust and adaptive learning despite extreme data heterogeneity, (2) effective preservation, weighting, and fusion of discriminative local features during server-side aggregation, and (3) richer, more comprehensive, and structurally informed identity modeling. Consequently, the RK module empowers the global model to integrate diverse and salient identity cues from multiple clients, significantly enhancing re-identification performance in challenging federated learning environments—while strictly preserving data privacy.
\\
\textbf{Dual-Branch Network.} RK employs a dual-branch design: a \textbf{Body Part Processing Branch} and a \textbf{Global Feature Processing Branch}. First, the PifPaf pose estimation model\cite{Kreiss_2019_CVPR}is used to segment pedestrian images into three main body regions—the head, the torso (including the left and right shoulders and hips), and the lower limbs (including the left and right knees and legs). The Body Part Processing Branch is used in the client-side model to process these body part images. Since the body part features learned by the Body Part Processing Branch vary depending on the local data distribution of the client, forcing the aggregation of the Body Part Processing Branch parameters would compromise the integrity of client-specific body part details. Therefore, this branch only runs during the training phase. During federated aggregation, only the parameters of the Global Feature Processing Branch are aggregated, and the server only distributes the parameters of the Global Feature Processing Branch. During training, the Body Part Processing Branch extracts domain-invariant knowledge into the main model (Global Feature Processing Branch), enhancing its generalization ability. The Global Feature Processing Branch processes raw full-image data. During the local training phase on the client side, we fuse the combined body part feature vectors extracted from the Body Part Processing Branch with the overall features output by the Global Feature Processing Branch, which constitutes the final output features of the client model.
\noindent
The final characteristics of each client output are defined as follows:
\begin{equation}
\theta_{\text {final }}=\partial \cdot \theta_{\text {global }}+(1-\partial) \cdot \theta_{\text {part }},
\end{equation}
where $\theta_{\text {global }}$ denotes the Global feature vector output by Global Feature Processing Branch, $\theta_{\text {part }}$ denotes the Combined local features output by the Body Part Processing Branch, $\partial$ denotes the weights assigned to the global features(default $\partial$ = 0.5) .
\noindent
The optimization objective of each client model is defined as follows:
\begin{equation}
L_{l o c}\left(x, y ; f_L\right)=L_{t r i}\left(\theta_{\text {L}}(x), y\right)+L_{c e}\left(f_{\text {L}}(x), y\right),
\end{equation}
where $\theta_{L}$ is the feature extractor part of local model $f_{L}$, $L_{\text{CE}}$ denotes the cross-entropy loss for part-level identification, $L_{\text{tri}}$ represents the triplet loss for metric learning.
\subsection{Federated Aggregation with Robust Knowledge
Selection}
\textbf{Motivation.} Although introducing Body Part Processing Branch enables each client to learn more refined and discriminative local features, there are still significant differences in the effectiveness of different clients in extracting such domain-invariant body-part-based knowledge. If the server simply aggregates all client models by averaging unified parameters, the unique contributions of clients that excel in capturing high-quality local details will be diluted by clients with poorer learning performance. This approach limits the generalization ability of the global model. To fully leverage the cross-domain invariant knowledge possessed by each client, we introduce KS(knowledge selection). This mechanism objectively evaluates and assigns higher aggregation weights to clients that excel in learning domain-invariant local features. The collaboration between RS and KS ensures that the global model benefits more from cross-domain invariant knowledge, ultimately enhancing overall performance and adaptability across diverse domains.
\\
\textbf{Directional Consistency Metric.} The direction consistency metric aims to assess the consistency between the update direction of each client's local model and the update direction of the global model through a bidirectional vector comparison framework. Specifically, this metric is calculated based on two types of feature differences: (1) the difference between the current global model and the aggregated features of the previous version of the global model, which reflects the update direction of the global model; (2) the difference between the features output by the client model before and after local training, which captures the local update direction of that client. The ratio of the norms of these two feature differences serves as a quantitative measure of directional consistency. The closer the ratio is to 1, the higher the alignment between the client's update direction and the global model's update direction. This ratio directly influences the dynamic weight adjustment mechanism during the aggregation process: clients with higher directional consistency receive higher aggregation weights, while those with lower directional consistency receive lower aggregation weights.

The directional consistency metric and the aggregation weight are mathematically defined as follows:
\begin{equation}
\mathcal{R}_k^{(t)}=\frac{\left\|\theta_g^{(t)}-\theta_g^{(t-1)}\right\|_2}{\left\|\theta_k^{(t+\Delta t)}-\theta_k^{(t)}\right\|_2},
\end{equation}
where $\theta_g^{(t)}$ denotes the global model parameters after the $t$-th round of aggregation, serving as the server-side global state to guide client initialization.The initial parameters after receiving the new global model are represented as $\theta_k^{(t)}$.Subsequently, the client performs $\Delta$$t$ rounds of optimization based on local data, yielding the updated  $\theta_k^{(t+\Delta t)}$,which reflect its personalized adjustment results.
\\
\textbf{Dynamic Weight Allocation.} Based on the direction consistency metric, our dynamic aggregation strategy can adaptively determine the contribution weight of each client in federated aggregation. Specifically, the weight assigned to each client is an exponentially decaying function of the deviation between its direction consistency ratio and the ideal value of 1, controlled by a temperature coefficient. This ensures that clients whose local updates are highly consistent with the global target receive higher aggregation weights, while clients with larger deviations are correspondingly assigned lower weights. Before aggregation, the obtained weights are normalized to ensure that the sum is 1. To further enhance stability, a protection mechanism freezes the weights of clients with very small local updates to prevent noise amplification, and applies historical smoothing to prevent the aggregated weights from undergoing drastic changes between different rounds.

The adaptive weighting mechanism and associated enhancement techniques are mathematically defined as follows:
\begin{equation}
\alpha_k^{(t)}=\frac{\exp \left(-\beta\left|1-\mathcal{R}_k^{(t)}\right|\right)}{\sum_{j=1}^K \exp \left(-\beta\left|1-\mathcal{R}_j^{(t)}\right|\right)},
\end{equation}
where $\alpha_k^{(t)}$ denotes the aggregation weight assigned to client $k$ at communication round $t$, $\beta$ is the temperature coefficient (default $\beta = 5.0$) controlling weight decay intensity.
\begin{equation}
\theta_g^{(t+1)}=\sum_{k=1}^K \alpha_k^{(t)} \cdot \theta_k^{(t+\Delta t)},
\end{equation}
where ($\theta_g^{(t+1)}$) denotes the global model parameters after the ($t+1$)th round of aggregation.
\begin{equation}
\tilde{\mathcal{R}}_k^{(t)}=\gamma \mathcal{R}_k^{(t)}+(1-\gamma) \tilde{\mathcal{R}}_k^{(t-1)},
\end{equation}
where $\gamma \in [0,1]$ denotes the smoothing factor (default $\gamma = 0.7$) for historical consistency regularization.
\begin{equation}
\alpha_k^{(t)}= \begin{cases}0 & \text { if }\left\|\theta_k^{(t+\Delta t)}-\theta_k^{(t)}\right\|_2<\varepsilon ,\\ \alpha_{k}^{(t)} & \text { otherwise ,}\end{cases}
\end{equation}
where $\varepsilon$ represents the update magnitude threshold (default $\varepsilon = 10^{-8}$) for confidence gating.
\begin{algorithm}[t]
\caption{\textbf{FedARKS}: Federated Aggregation via Robust and Discriminative Knowledge Selection}
\label{alg:FedARKS}
\begin{algorithmic}[1]
\REQUIRE Total rounds $T$, client set $\mathcal{C}$, local epochs $E$, temperature $\beta$, threshold $\varepsilon$
\ENSURE Optimized global model $\theta_g$

\STATE Initialize global model parameters $\theta_g^{(0)}$
\FOR{each round $t = 1,2,\ldots,T$}
    \STATE \textit{// Server broadcasts the global model}
    \STATE Send $\theta_g^{(t)}$ to all clients $k \in \mathcal{C}$
    \FOR{each client $k \in \mathcal{C}$ \textbf{in parallel}}
        \STATE Initialize local model $\theta_k \gets \theta_g^{(t)}$
        \FOR{epoch = 1 to $E$}
            \STATE \textit{// Local dual-branch RK training:}
            \STATE \quad Update $\theta_k$ using RK network with
            \STATE \quad $\bullet$ \textbf{Global branch}: learns holistic features for aggregation
            \STATE \quad $\bullet$ \textbf{Part branch}: extracts domain-invariant features to regularize the global branch
        \ENDFOR
        \STATE Compute reliability score $\mathcal{R}_k$ for the global branch
        \STATE Upload $(\theta_k, \mathcal{R}_k)$ to server
    \ENDFOR
    \STATE \textit{// Adaptive knowledge selection and aggregation}
    \FOR{each client $k \in \mathcal{C}$}
        \STATE $w_k \gets \exp(-\beta |1-\mathcal{R}_k|)$ if $\mathcal{R}_k > \varepsilon$; else $w_k \gets 0$
    \ENDFOR
    \STATE Normalize weights: $\alpha_k = \frac{w_k}{\sum_j w_j}$
    \STATE Update $\theta_g^{(t+1)} \gets \sum_{k \in \mathcal{C}} \alpha_k \cdot \theta_k$
\ENDFOR
\STATE \textbf{Return} $\theta_g^{(T)}$
\end{algorithmic}
\end{algorithm}
\section{Experiments}
\subsection{Experimental Setup}
\textbf{Datasets.} We evaluate our method on four person re-identification (ReID) datasets:
\begin{itemize}
    \item \textbf{CUHK02} \cite{li2013locally}: Contains 1,816 identities captured by 2 camera views, with 7,264 annotated bounding boxes.
    
    \item \textbf{CUHK03} \cite{li2014deepreid}: Includes 1,467 identities from 2 camera views, providing 14,096 detected or manually labeled bounding boxes.
    
    \item \textbf{Market1501} \cite{zheng2015scalable}: Comprises 32,668 bounding boxes of 1,501 identities captured by 6 disjoint cameras.
    
    \item \textbf{MSMT17} \cite{wei2018person}: A large-scale dataset with 126,441 bounding boxes of 4,101 identities from 15 cameras, featuring complex lighting and weather variations.
\end{itemize}
\noindent
\textbf{Model. } For the person re-identification tasks on all datasets, we employ a ResNet50 backbone. During federated training, each client appends a client-specific classifier head where the output dimension equals the number of identities in the client's local dataset.For the ViT backbone, we used a batch size of 16 during training.
\begin{table*}[t]
\centering
\renewcommand{\arraystretch}{1.15}
\small
\begin{tabular}{c|c|c|cc|cc|cc|cc}
\Xhline{1.5pt}
\multirow{2}{*}{\textbf{Category}} & \multirow{2}{*}{\textbf{Method}} & \multirow{2}{*}{\textbf{Venue}} &
\multicolumn{2}{c|}{\textbf{MS+C2+C3→M}} &
\multicolumn{2}{c|}{\textbf{M+C2+C3→MS}} &
\multicolumn{2}{c|}{\textbf{MS+C2+M→C3}} &
\multicolumn{2}{c}{\textbf{SAvg}} \\
\cline{4-11}
 & & & \textbf{mAP} & \textbf{R1} & \textbf{mAP} & \textbf{R1} & \textbf{mAP} & \textbf{R1} & \textbf{mAP} & \textbf{R1} \\
\hline
\multicolumn{11}{l}{\textbf{Backbone: ResNet50}} \\
\hline
\multirow{3}{*}{Federated Learning} 
    & SCAFFOLD      & ICML’20   & 26.0 & 50.5 & 5.3 & 15.8 & 22.9 & 26.0 & 18.1 & 30.8 \\
    & MOON          & CVPR’21   & 26.8 & 51.1 & 4.8 & 14.5 & 20.9 & 22.5 & 17.5 & 29.4 \\
    & FedProx       & MLSys’20  & 29.3 & 53.8 & 5.8 & 17.4 & 19.1 & 17.7 & 18.1 & 29.6 \\
\hline
\multirow{3}{*}{Domain Generalization} 
    & SNR           & CVPR’20   & 32.7 & 59.4 & 5.1 & 15.3 & 28.5 & 30.0 & 22.1 & 34.9 \\
    & MixStyle      & ICLR’21   & 31.2 & 53.5 & 5.5 & 16.0 & 28.6 & 31.5 & 21.8 & 33.7 \\
    & CrossStyle    & ICCV’21   & 35.5 & 59.6 & 4.6 & 14.0 & 27.8 & 28.0 & 22.6 & 33.9 \\
\hline
\multirow{2}{*}{Fed ReID}
    & FedPav        & MM’20     & 25.4 & 49.4 & 5.2 & 15.5 & 22.5 & 24.3 & 17.7 & 29.7 \\
    & FedReID       & AAAI’21   & 30.1 & 53.7 & 4.5 & 13.7 & 26.4 & 26.5 & 20.3 & 31.3 \\
\hline
\multirow{4}{*}{FedDG-ReID}
    & DACS          & AAAI’24   & 35.5 & 60.3 & 9.0 & 26.7 & 24.6 & 24.4 & 23.0 & 37.1 \\
    & SSCU          & MM’25   & 39.5 & 66.4 & 11.9 & 32.3 & 32.8 & 34.1 & 28.1 & 44.3 \\
    & FedSupWA          & AAAI’26   & 40.3 & 67.5 & 12.4 & 33.1 & 33.1 & 35.6 &29.2 & 45.3 \\
    & \textbf{FedARKS} & -      & \textbf{41.5} & \textbf{68.2} & \textbf{12.6} & \textbf{33.7} & \textbf{34.8} & \textbf{35.9} & \textbf{29.6} & \textbf{45.9} \\
\hline
\multicolumn{11}{l}{\textbf{Backbone: ViT}} \\
\hline
\multirow{1}{*}{Fed ReID}
    & FedPav        & MM’20     & 32.5 & 56.0 & 13.6 & 32.6 & 25.1 & 24.0 & 23.7 & 37.5 \\
\hline
\multirow{1}{*}{DG}
    & CrossStyle    & ICCV’21   & 30.9 & 55.1 & 16.8 & 39.9 & 19.2 & 18.2 & 22.3 & 37.7 \\
\hline
\multirow{2}{*}{FedDG-ReID}
    & DACS          & AAAI’24   & 38.3 & 63.7 & 17.3 & 39.9 & 29.1 & 28.8 & 28.2 & 44.1 \\
    & \textbf{FedARKS} & -      & \textbf{43.6} & \textbf{68.4} & \textbf{18.2} & \textbf{40.6} & \textbf{33.0} & \textbf{33.3} & \textbf{31.6} & \textbf{47.4} \\
\Xhline{1.5pt}
\end{tabular}
\caption{Comparison of methods with different backbones and target unseen domains. M:Market1501, C2:CUHK02, C3:CUHK03, MS:MSMT17,R1: Rank-1. SAvg columns report the average mAP and average R1 across the three transfer tasks,highlighting model generalization ability.}
\label{table1}
\end{table*}
\subsection{Comparison to State-of-the-Arts}
As shown in Tab.\ref{table1}, our method consistently outperforms all baselines across different datasets and backbones. Under both ResNet50 and ViT, it achieves the best results, surpassing DACS \cite{yang2024diversity} in every evaluation scenario. The gains are particularly evident on challenging datasets such as MSMT17 and CUHK03, highlighting our approach’s strong robustness to both backbone variations and unseen domains. For example, in the M+C2+C3→MS configuration with ViT, our method achieves 18.2\% mAP and 40.6\% R1, demonstrating excellent scalability to large datasets. Similarly, in MS+C2+M→C3, it maintains 31.5\% mAP and 34.7\% R1, reflecting high adaptability even in small-scale target domains. Our approach performs strongly with both powerful and lightweight backbone architectures, ensuring practical versatility. Results on source domains (see Tab.\ref{table2}) further validate the effectiveness of our method. FedARKS achieves state-of-the-art performance on all benchmark datasets while adhering to federated learning constraints. Specifically, on Market1501, it sets new SOTA results of 73.5\% mAP and 89.4\% Rank-1 accuracy (R1), exceeding DACS by \textbf{$\uparrow$~1.4\%} mAP and \textbf{$\uparrow$~1.2\%} R1. Gains are observed on CUHK02, with 86.8\% mAP and 86.5\% R1 – improvements of \textbf{$\uparrow$~2.3\%} mAP and \textbf{$\uparrow$~3.1\%} R1 over DACS. For the challenging CUHK03 dataset, FedARKS attains 54.5\% mAP and 56.8\% R1, surpassing DACS by \textbf{$\uparrow$~7.1\%} mAP and \textbf{$\uparrow$~6.7\%} R1. It outperforms the best prior federated method (FedProx) by over 30\% in mAP on CUHK03. These results demonstrate FedARKS’s superior cross-domain generalization, architectural adaptability, and scalability for real-world person ReID tasks. Moreover, SAvg results clearly further demonstrate superiority of our method across all evaluated settings.
\begin{table}[ht]
\centering
\renewcommand{\arraystretch}{1.25}
\small
\setlength{\tabcolsep}{4pt}
\begin{tabular}{@{}>{\centering\arraybackslash}p{1.5cm}*{6}{c}@{}}
\toprule
\multirow{2}{*}{\textbf{Method}} & 
\multicolumn{2}{c}{\textbf{Market1501}} & 
\multicolumn{2}{c}{\textbf{CUHK02}} & 
\multicolumn{2}{c}{\textbf{CUHK03}} \\
\cmidrule(lr){2-3} \cmidrule(lr){4-5} \cmidrule(l){6-7}
 & \textbf{mAP} & \textbf{R1} & \textbf{mAP} & \textbf{R1} & \textbf{mAP} & \textbf{R1} \\
\midrule
FedProx & 61.0 & 80.4 & 66.8 & 65.5 & 24.2 & 23.9 \\
FedPav & 53.9 & 76.0 & 59.7 & 56.3 & 19.6 & 19.6 \\
FedReID & 71.8 & 87.6 & 82.9 & 82.8 & 44.0 & 44.9 \\
DACS & 72.1 & 88.2 & 84.5 & 83.4 & 47.4 & 50.1 \\
\textbf{FedARKS} & \textbf{73.5} & \textbf{89.4} & \textbf{86.8} & \textbf{86.5} & \textbf{54.5} & \textbf{56.8} \\
\bottomrule
\end{tabular}
\caption{Source domain performance comparison. Training domains: Market1501 + CUHK02 + CUHK03; Test domains: Market1501, CUHK02, and CUHK03.}
\label{table2}
\end{table}

\begin{table}[t]
\centering
\renewcommand{\arraystretch}{1.25}
\small
\begin{tabular}{c|c|c|c}
\Xhline{1.5pt}

\multicolumn{1}{c|}{} & \multicolumn{1}{c|}{\textbf{MS+C2}} & \multicolumn{1}{c|}{\textbf{C3+C2}} & \multicolumn{1}{c}{\textbf{MS+C2}} \\

\multicolumn{1}{c|}{\multirow{-2}{*}{\textbf{Attributes}}} & \multicolumn{1}{c|}{\textbf{+C3→M}} & \multicolumn{1}{c|}{\textbf{+M→MS}} & \multicolumn{1}{c}{\textbf{+M→C3}} \\
\hline

RK\hspace{0.6em}KS & mAP\hspace{0.8em}R1 & mAP\hspace{0.8em}R1 & mAP\hspace{0.8em}R1 \\
\hline
\hline
$\times$\hspace{1.3em}$\times$ & 
40.3\hspace{0.8em}67.5 & 12.4\hspace{0.8em}33.1 & 33.1\hspace{0.8em}35.6 \\

\checkmark \hspace{1.3em}$\times$ & 40.8\hspace{0.8em}67.9 & 
12.5\hspace{0.8em}33.4 & 33.6\hspace{0.8em}35.8\\

$\times$\hspace{1.3em}\checkmark & 
41.0\hspace{0.8em}68.0& 
12.5\hspace{0.8em}33.9 & 34.0\hspace{0.8em}35.8 \\

\checkmark\hspace{1.3em}\checkmark & 
\textbf{41.5}\hspace{0.8em}\textbf{68.2}& 
\textbf{12.6}\hspace{0.8em}\textbf{33.7}& 
\textbf{34.8}\hspace{0.8em}\textbf{35.9}\\
\Xhline{1.5pt}
\end{tabular}
\caption{Ablation study on Robust Knowledge(RK) and Knowledge Selection(KS).}
\label{table3}
\end{table}
\subsection{Ablation Study}
In order to better understand the impact of \textbf{RK} and \textbf{KS} on the FedDG-ReID results, we gradually incorporated them into the training processes of MS+C2+C3 → M, MS+C2+M → C3, and M+C2+C3 → MS for ablation studies.We compare four different training schemes in Tab.\ref{table3}. and summarize the following conclusions:
\\
\textbf{Synergistic Effect of Dual Components}: Integrating both RK and KS yields the most substantial improvements across all tasks, demonstrating strong complementarity. For example, in the MS+C2+C3→M setting, their combination boosts mAP by \textbf{$\uparrow$~1.2\%} and R1 by \textbf{$\uparrow$~0.7\%}, while in MS+C2+M→C3, the R1 gain reaches \textbf{$\uparrow$~0.3\%}.
\\
\textbf{RK's Feature Discrimination Contribution}: RK is most effective in transfers between similar domains, notably improving R1 in MS+C2+C3→M by \textbf{$\uparrow$~0.5\%}. Its combination with KS provides additional benefits, as seen in the incremental mAP gain in the M+C2+C3→MS task.
\\
\textbf{KS's Cross-Domain Generalization Strength}: KS excels in challenging domain shifts, achieving larger improvements than RK, especially in MS+C2+M→C3 (\textbf{$\uparrow$~0.9\%} mAP and \textbf{$\uparrow$~0.2\%} R1 over baseline), establishing KS as the key to cross-domain adaptation.
\\
\textbf{Complementary Properties:} Tab.\ref{table3} shows that using both RK and KS together delivers greater mAP and R1 improvements than either alone—e.g., in MS+C2+M→C3, mAP increases from 33.6\% (RK only) or 34.0\% (KS only) to 34.8\% (both), and R1 from 35.8\% (RK) or 35.8\% (KS) to 35.9\% (both). This clearly confirms their complementarity and the effectiveness of joint use for enhancing federated cross-domain ReID performance.

\begin{figure}[t]
\centering
\includegraphics[width=0.86\columnwidth]{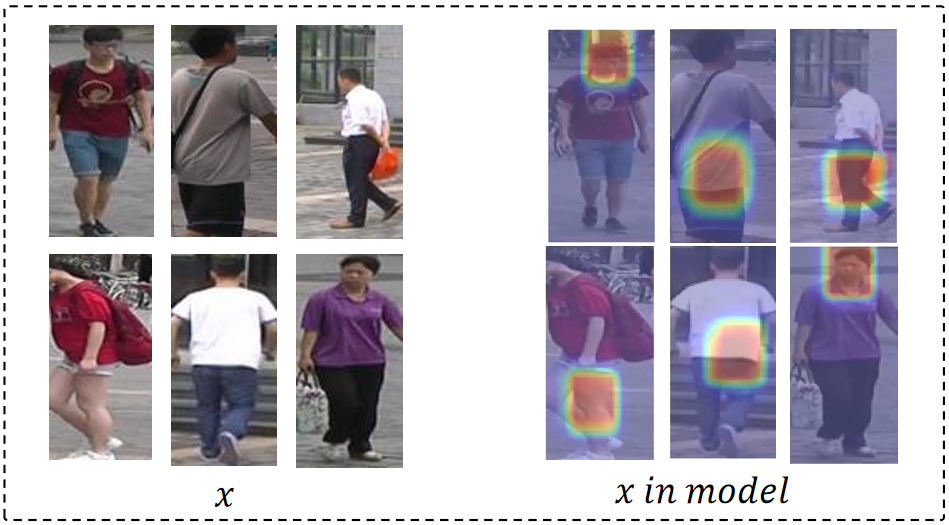}
\caption{Dynamic Attention Distribution Heatmap Comparison Across Diverse Scenarios.}
\label{fig3}
\end{figure}
\begin{figure}[t]
\centering
\includegraphics[width=0.9\columnwidth]{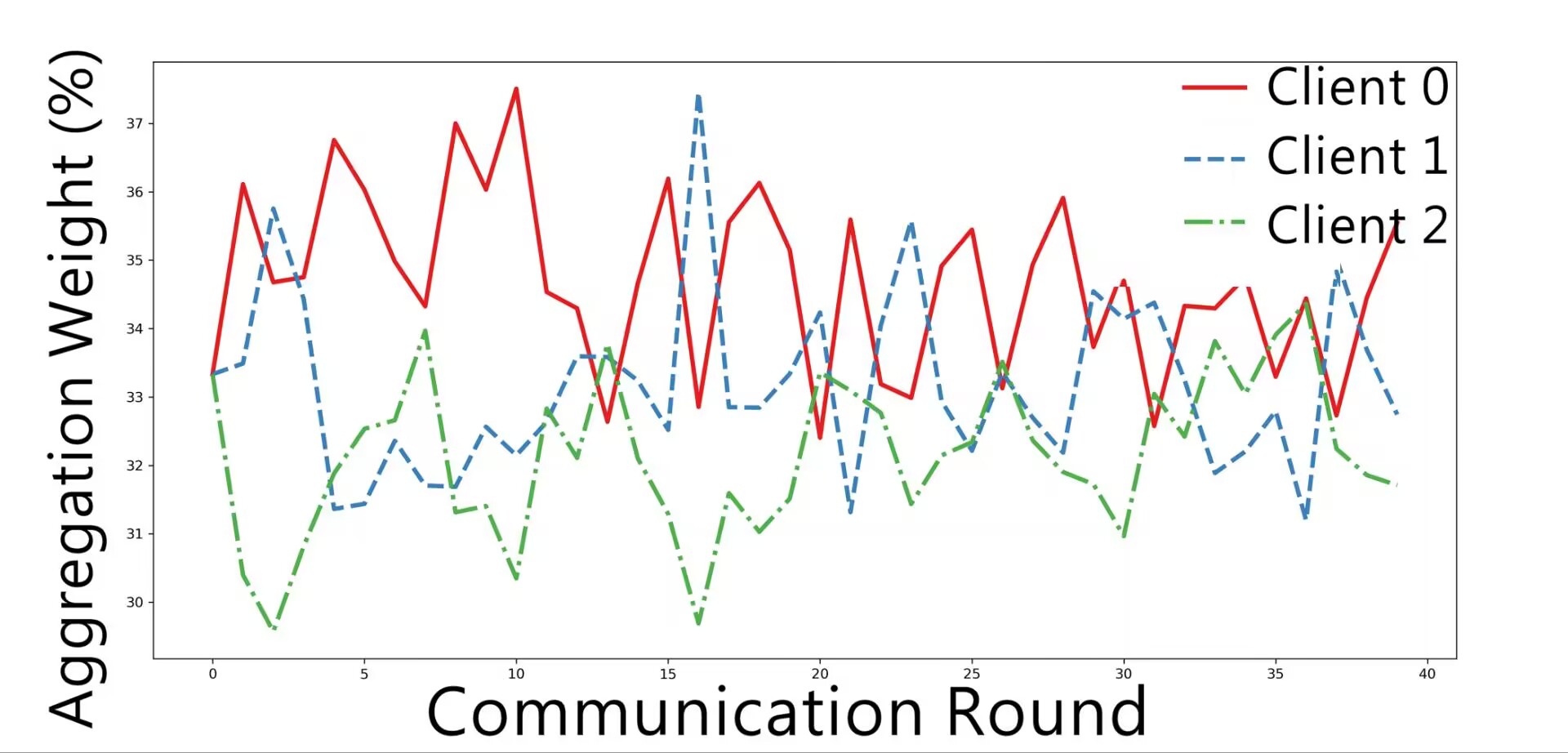}
\caption{Dynamic client weight distribution across 40 training epochs in FedARKS.}
\label{fig4}
\end{figure}
\subsection{Visualization} To gain a deeper understanding of our RK mechanism’s practical effects, we present comprehensive attention heatmaps comparing network focus patterns in a variety of real-world and challenging scenarios. In standard viewpoints, the RK module adaptively highlights the head and face, which are the most identifiable features for person ReID. In complex conditions—such as rare camera perspectives or cluttered backgrounds—the heatmaps show the model shifting focus to other discriminative regions, such as the torso or lower limbs. This dynamic adaptation proves to be crucial: for example, when the face is occluded or indistinct, unique clothing textures or accessories on the lower body become primary sources of cross-client invariance. Thus, the RK mechanism greatly enhances clients’ ability to mine and use contextually relevant identity features, while promoting aggregation of representations that are both semantically meaningful and robust to environmental or camera-view changes. This effectively addresses poor generalization under non-IID conditions, ensuring the learned global model captures ensembles of locally optimal, scene-adaptive features critical for robust cross-domain ReID. Additionally, Fig.\ref{fig4} vividly shows the dynamic evolution of aggregated weights for each client over 40 federated training epochs. The curves indicate every client’s contribution is continuously adjusted based on its local updates and, as training converges, gradually stabilizes. This means the KS mechanism robustly integrates clients with different key detail learning capabilities, even with severely non-IID data. Dynamic weighting effectively prevents low-quality or mismatched client updates from causing negative transfer and helps the global model better absorb valuable cross-domain invariant knowledge. Overall, these visualization results clearly demonstrate our RK and KS mechanisms work synergistically for reliable feature learning in challenging joint ReID tasks.

\section{Conclusion}
This paper proposes a novel federated aggregation via robust and discriminative knowledge selection and integration (FedARKS) to address the inherent challenges of FedDG-ReID. Specifically, our RK architecture combines local body part feature learning with global features, enabling client models to learn key local details and richer context. KS dynamically adjusts aggregation weights by evaluating directional consistency between client updates and global optimization trajectories, ensuring robust semantic consistency and fine-grained alignment in the federated feature space. Extensive experiments show that our method demonstrates significant robustness on popular backbones like ResNet50 and ViT, further validating its scalability and effectiveness for privacy-preserving person ReID in distributed learning scenarios with heterogeneous clients.

\section{Acknowledgments}
This work was supported by the Natural Science Foundation of China (62376201), Hubei Provincial Science \& Technology Talent Enterprise Services Program (2025DJB059), and Hubei Provincial Special Fund for Central-Guided Local S\&T Development (2025CSA017).

\bigskip

\bibliography{AAAI_camera_ready/aaai2026}

\end{document}